\documentclass[10pt,twocolumn,letterpaper]{article}

\usepackage{cvpr}
\usepackage{times}
\usepackage{epsfig}
\usepackage{graphicx}
\usepackage{amsmath}
\usepackage{amssymb}
\usepackage{soul}
\usepackage[utf8]{inputenc}
\usepackage{graphicx}
\usepackage{amsmath}
\usepackage{amssymb}
\usepackage{color}
\usepackage{multirow}
\usepackage{diagbox}
\usepackage{rotating}
\usepackage{subcaption}
\usepackage{booktabs}
\usepackage{array}
\usepackage{colortbl}
\usepackage{enumitem}
\usepackage{url}

\usepackage[pagebackref=true,breaklinks=true,letterpaper=true,colorlinks,bookmarks=false]{hyperref}

\newcolumntype{x}[1]{>{\centering\arraybackslash\hspace{0pt}}p{#1}}

\usepackage{makecell}
\usepackage{hhline}
\usepackage{cellspace}
\usepackage{mathtools}
\DeclarePairedDelimiterX\set[1]\{\}{\nonscript\,#1\nonscript\,}
\usepackage{pifont}

\newcommand{\PreserveBackslash}[1]{\let\temp=\\#1\let\\=\temp}
\newcolumntype{C}[1]{>{\PreserveBackslash\centering}p{#1}}
\newcolumntype{R}[1]{>{\PreserveBackslash\raggedleft}p{#1}}
\newcolumntype{L}[1]{>{\PreserveBackslash\raggedright}p{#1}}
\def\S4Net{{S4Net}}
\def\sArt{state-of-the-art~}
\definecolor{gray1}{rgb}{.8,.8,.8}

\newcommand{\figref}[1]{Fig. \ref{#1}}
\newcommand{\tabref}[1]{Tab. \ref{#1}}

\newcommand{\secref}[1]{Sec. \ref{#1}}

\renewcommand{\arraystretch}{1}
\renewcommand{\tabcolsep}{.5mm}

\newcommand{\myPara}[1]{\vspace{-.04in}\paragraph{#1}}
\newcommand{\CaptSpace}{\vspace{-.08in}}

\graphicspath{{./figs/}}
\newcommand{\addFig}[1]{}
\newcommand{\addFigs}[1]{}

\cvprfinalcopy 

\def\cvprPaperID{3132} 

\ifcvprfinal\fi
\begin{document}

\title{S4Net: Single Stage Salient-Instance Segmentation}

\author{Ruochen Fan$^{1,2}$\quad Ming-Ming Cheng$^3$ \quad Qibin Hou$^3$ \quad 
    Tai-Jiang Mu$^{1,2}$\quad     Jingdong Wang$^4$\quad   Shi-Min Hu$^{1,2}$\\
$^1$Tsinghua University \quad $^2$BNRist \quad $^3$Nankai University  \quad $^4$MSRA \\
}

\maketitle

\begin{abstract}
We consider an interesting problem---salient instance segmentation
in this paper.
Other than producing bounding boxes,
our network also outputs high-quality instance-level segments.
%
Taking into account the category-independent property of each target,
we design a single stage salient instance segmentation framework,
with a novel segmentation branch.
Our new branch regards not only local context
inside each detection window but also its surrounding context,
enabling us to distinguish the instances in the same scope even with obstruction.
Our network is end-to-end trainable and runs at a fast speed
(40 fps when processing an image with resolution $320 \times 320$).
We evaluate our approach on a public available benchmark
and show that it outperforms other alternative solutions.
We also provide a thorough analysis of the design choices
to help readers better understand the functions of each part of our network.
The source code can be found at \url{https://github.com/RuochenFan/S4Net}.
\end{abstract}


\maketitle

\section{Introduction}


Rather than recognizing all the objects in a scene,
we human only care about a small set of interesting
objects/instances \cite{li2002rapid}.
A recent experiment \cite{elazary2008interesting} demonstrates that
interesting objects are often
visually salient, reflecting the importance of detecting salient objects.
Locating objects of interest is also essential
for a wide range of computer graphics and computer vision applications.
Such a capability allows many modern applications
(\eg image manipulation/editing
\cite{cheng2010repfinder,wu2010resizing,tog09Sketch2Photo}
and robotic perception \cite{wu2014hierarchical})
to provide initial regions that might be of interest to users
or robots so that they can directly proceed to
image editing or scene understanding.
Like \cite{li2017instance}, in this paper,
we are also interested in detecting \emph{salient instances}
given an input image.
Similar to salient object detection, salient instance segmentation aims at
detecting the most distinctive objects in a scene, but differently it also
identifies each individual instance, \ie outputting
an accurate segment for each instance and assigning it a unique label
(see \figref{fig:demo}).


\newcommand{\AddFig}[1]{\includegraphics[width=0.32\linewidth]{Demos/#1.jpg}}

\begin{figure}[t]
  \centering
  \renewcommand{\tabcolsep}{1pt}
  \AddFig{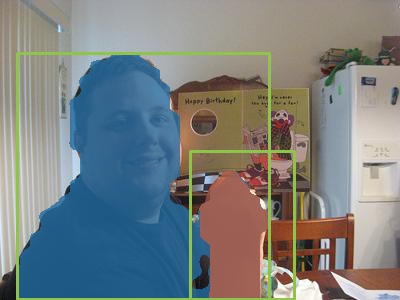}
  \AddFig{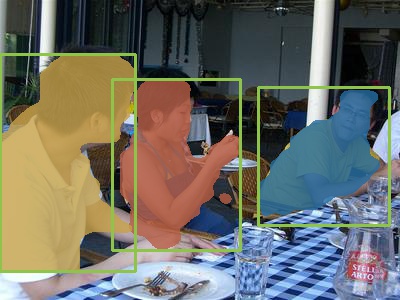}
  \AddFig{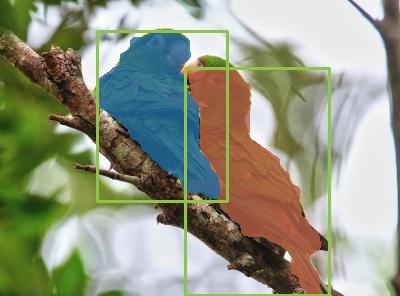} \\
  \CaptSpace
  \caption{Illustrative examples produced by our approach,
  	which detects and segments salient instances regardless of their
  	semantic categories.
  	Each category-agnostic salient instance is illustrated with a unique color.
 }\label{fig:demo}
\end{figure}

Cognitive psychology \cite{wolfe2004attributes,desimone1995neural}
and neurobiology \cite{mannan2009role} research suggested that
human cortical cells may be \emph{hard wired} to
preferentially respond to \emph{high contrast} stimulus,
\ie feature separation between foreground and background regions
plays a central role in salient object perception
\cite{itti1998model,itti2001computational}.
Effectively modeling the foreground/background separation
using local and global contrast \cite{cheng2015global,jiang2013salient},
background prior \cite{zhu2014saliency},
and Gaussian mixture color model (as in GrabCut \cite{rother2004grabcut}),
\etc, has been proven to be useful in a variety of
traditional salient-object/figure-ground segmentation tasks.
Recently, convolutional neural networks (CNNs) are becoming the dominant
methods in nearly all closely related tasks, \eg
salient object detection \cite{hou2016deeply,li2016deep,wang2015deep},
semantic instance segmentation \cite{dai2015convolutional,
hariharan2014simultaneous,hariharan2015hypercolumns},
and generic object detection \cite{girshick2014rich,ren2017faster,dai2016r}.
While these CNN-based methods have achieved remarkable
success by learning powerful multi-level feature representations
for capturing different  abstracted
\emph{appearance variations} of the target categories,
they often ignore the important \emph{feature separation ability
between the target objects and their nearby background}.

Existing CNN-based instance segmentation methods
use either RoIPooling \cite{fastRcnn15,he2015spatial},
or RoIWarp \cite{dai2016instance}, or RoIAlign \cite{he2017mask}
to capture the \emph{feature information inside the bounding boxes}.
In contrast,
we propose a region feature extraction layer, namely \emph{RoIMasking},
to explicitly \emph{incorporate foreground/background separation} for
improving salient instance segmentation.
Similar to the figure-ground segmentation
method, GrabCut \cite{rother2004grabcut},
we explicitly mark the region surrounding the object proposals
as the initial background,
and explore the foreground/background feature separations for
salient instance segmentation in our segmentation branch.
More specifically, we flip the signs of the feature values
surrounding the proposals.
%
%
The RoIMasking layer based segmentation branch is then integrated
to the efficient single-stage object detector FPN \cite{lin2017feature},
for detecting the pixel-wise segment of each salient instance.
Interestingly, our RoIMasking scheme is quantization-free
and scale-preserving, allowing more detailed
information to be successfully detected.
Furthermore, our model is end-to-end trainable and runs at 40fps
on a single GPU when processing a $320 \times 320$ image.

For verification in the context of the killer application,
we apply our salient instance detector to the popular
weakly-supervised semantic segmentation task.
As done in \cite{wei2016stc,hou2016mining},
we use the detected salient instances
on the tiny ImageNet dataset \cite{russakovsky2015imagenet,hou2016mining}
as heuristics to train the famous semantic segmentation networks.
We evaluate the results on the popular PASCAL VOC 2012 semantic segmentation
benchmark \cite{everingham2015pascal} and show that
our results outperform \sArt methods \cite{wei2016stc,hou2016mining} that leverage
traditional salient object cues \cite{jiang2013salient,hou2016deeply}
by a large margin.


To sum up, the contributions of this paper are:
\setlist{nolistsep}
\begin{itemize}[noitemsep]
  \item We propose an end-to-end single-shot salient instance
  	 segmentation framework,
  	 which not only achieves the \sArt performance
  	 but also runs in real time.
  \item We design a new RoIMasking layer which 
  	models feature separation between target objects and its nearby
  	background for high-quality segmentation.
\end{itemize}

\begin{figure*}[ht!]
  \centering
  \includegraphics[width=\linewidth]{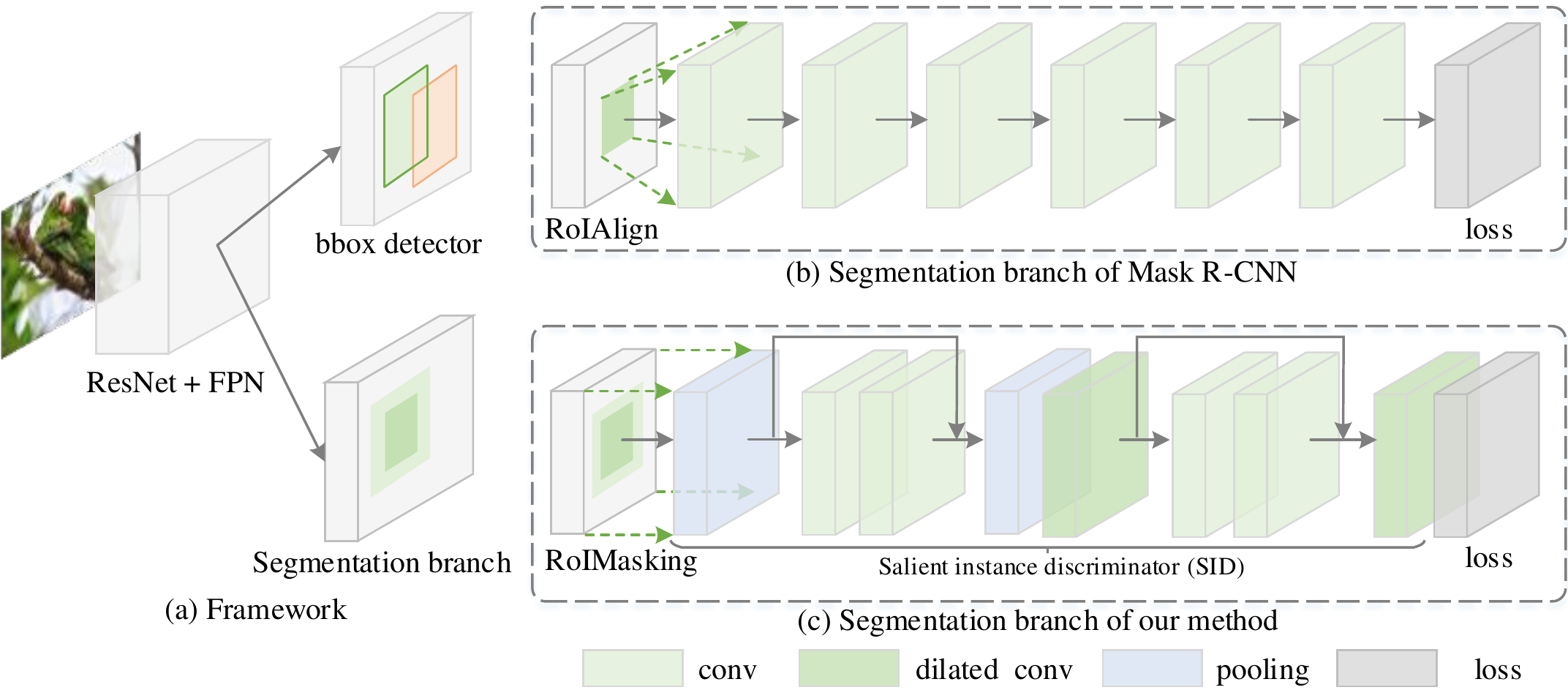}\\
  \CaptSpace
  \caption{The pipeline of the proposed method.
  	(a) A brief illustration of our framework.
  	For convenience, we do not show the details of the backbone we adopt.
	Readers may refer to \cite{lin2017feature} for more information.
	(b) The segmentation branch proposed in Mask R-CNN \cite{he2017mask},
	which is composed of a stack of consecutive convolutional layers.
	(c) Our proposed segmentation branch which further enlarges the size of
	the receptive field but with the same number parameters as in (b).
  }\label{fig:arch}
\end{figure*}

\section{Related Works}

Salient instance segmentation is relatively a new task.
Some seminal methods have recently been proposed by Zheng \etal
\cite{Zhang_2016_CVPR} to find salient objects at bounding box level.
However, this method misses the important segmentation information,
which is essential for applications such as
image editing \cite{cheng2010repfinder,tog09Sketch2Photo}
and weakly supervised segmentation \cite{wei2016stc}.
Li \etal \cite{li2017instance} formally define
the salient instance segmentation problem as jointly identifying
salient regions as well as individual object instances.
They also proposed an MSRNet \cite{li2017instance} framework for
instance-level salient object segmentation.
However, this method was excessively reliant on the quality of
the pre-computed edge maps (\eg MCG \cite{pont2017multiscale})
and produced sub-optimal results for
complicated real-world scenes (see also \secref{sec:experiments}).
%
Salient instance segmentation is closely related to
three major computer vision tasks:
salient object detection, object detection, and
semantic instance segmentation.

\subsection{Salient Object Detection}

Salient object detection aims at jointly detecting the most
distinguished objects and segmenting them out from a given scene.
Early salient object detection methods mostly depended on
either global or local contrast cues
\cite{qi2015saliencyrank, cheng2015global,borji2015salient,jiang2013salient}.
They designed various hand-crafted features
(e.g., color histogram and textures) for each region
\cite{achanta2012slic,felzenszwalb2004efficient,shi2000normalized} and
fused these features in either manual-designed \cite{cheng2015global}
or learning-based manners \cite{WangDRFI2017}.
Because of their weak ability to preserve the integrity of
salient instances and the instability of hand-crafted features,
these methods were gradually taken place by later CNN-based
data-driven methods \cite{hou2016deeply,li2016deep,
wang2015deep,zhao2015saliency,lee2016deep,li2015visual,li2017instance}.
The key problems of these salient object detection methods
when applied to salient instance segmentation task are two-fold.
First, the integrity of the salient objects
is difficult to be preserved because the distinguished regions
might be parts of the interesting instances.
Second, salient object detection is a binary problem
and hence cannot be competent to instance-level segmentation.

\subsection{Object Detection}

The goal of object detection is to produce all the bounding boxes
for semantic categories.
Earlier work mostly relied on hand-engineered features
(\eg SIFT \cite{lowe2004distinctive}, SURF \cite{bay2008speeded},
and HOG \cite{dalal2005histograms}).
They built different types of image pyramids to
leverage more information across scales.
Recently, the emergence of CNNs greatly promoted
the development of object detectors.
For example, R-CNN \cite{girshick2014rich} and
OverFeat \cite{sermanet2014overfeat} regarded CNNs
as sliding window detectors for extracting high-level semantic information.
Given a stack of pre-computed proposals
\cite{uijlings2013selective,cheng2014bing},
these methods computed its feature vectors for each proposal
using CNNs and then fed the features into a classifier.
Later work \cite{he2015spatial,fastRcnn15} took as inputs
the entire images and applied region-based detectors to feature maps,
substantially accelerating the running speed.
Faster R-CNN \cite{ren2017faster} broke through the limitation
of using pre-computed proposals by introducing
a region proposal network (RPN) into CNNs.
In this way, the whole network could be trained end-to-end,
offering a better trade-off between
accuracy and speed compared to its previous work.
However, all the methods discussed above aim at outputting reliable
object bounding boxes rather than instance segments.

\subsection{Semantic Instance Segmentation.}

Earlier semantic instance segmentation methods
\cite{dai2015convolutional,hariharan2014simultaneous,
hariharan2015hypercolumns,pinheiro2015learning} were mostly based
on segment proposals generated by segmentation methods
\cite{uijlings2013selective,pont2017multiscale,arbelaez2011contour}.
In \cite{dai2016instance},
Dai \etal predicted segmentation proposals by leveraging a multi-stage
cascade to refine rectangle regions from bounding box proposals gradually.
Li \etal \cite{li2016fully} proposed to integrate the
segment proposal network into an object detection network.
More recently, He \etal implemented a Mask R-CNN framework,
extending the Faster R-CNN \cite{ren2017faster}
architecture by introducing a segmentation branch.
Albeit more and more fascinating results,
these methods are not suitable for our task for two reasons.
First, not all the categories and objects  are salient.
Second, the semantic instances all belong to
a pre-defined category collection,
missing the important ability to deal with unknown categories,
\ie \emph{class-agnostic salient instances}.

%

\section{S4Net}

The design choices of our method are based on the application requirements
of high-quality salient instance segmentation in real time.
We design an end-to-end single-shot salient instance segmentation framework
---\textit{S4Net},
which is built upon the top of the \sArt single-shot object detector
for efficiency consideration.

\begin{figure}[t]
  \centering
  \renewcommand{\tabcolsep}{0.5pt}
  \includegraphics[width=\linewidth]{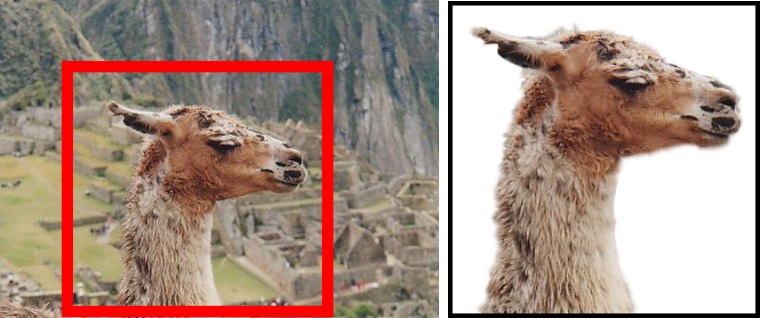} \\
  \CaptSpace
  \caption{An example of interactive figure-ground segmentation
  	using GrabCut \cite{rother2004grabcut}.
  }\label{fig:GrabCut}
\end{figure}

\subsection{Observation}\label{sec:observation}

Recent instance-level semantic segmentation methods \cite{he2017mask,li2016fully}
have shown the strong ability to segmenting semantic instances,
using RoIWarp \cite{dai2016instance}, or RoIAlign \cite{he2017mask}.
However, 
the segmentation branches of these methods \emph{only
focus on the features inside the proposals} to describe the
appearance variations of the target instances themselves,
lacking the ability to distinguish different instances.

Before CNN-based methods became popular,
utilizing feature separation between foreground and background has
been the dominant mechanism in similar tasks such as
salient object detection \cite{cheng2015global,jiang2013salient,zhu2014saliency}
and figure-ground segmentation \cite{rother2004grabcut}.
The ability of effectively modeling the foreground-background feature
separation is so powerful that these methods
\cite{cheng2015global,rother2004grabcut,zhu2014saliency} could
achieve \emph{remarkable success by utilizing such feature separation
in the target image alone,
without any additional information by training on many images}.
An example is shown in \figref{fig:GrabCut}.
Users only need to draw a rectangle region (shown in red)
around the target object.
The GrabCut method \cite{rother2004grabcut} initializes
the foreground/background color models, \ie Gaussian Mixture Models (GMM),
using image pixels inside/outside the rectangle region respectively.
Amazing segmentation results could be achieved,
without learning from other training images.
Notice that the color of some target object regions in this image
is very similar to certain background regions (\ie the houses).
However, the GMM color model effectively captures the slight color difference
(indistinguishable to human eyes) in this specific image.
Such slight color difference only exists in this specific image,
and could be impossible to learn from training examples of similar scenes.
It means that many training examples for color feature modelling will not only
be expensive to collect but also be less useful for dealing with such a situation.

Unfortunately, the ability of involving such powerful
foreground-background feature separation has been missing in
existing CNN-based segmentation methods.
Motivated by this, we propose to explicitly leverage more features
corresponding to the background area to help make the salient instances
more prominent as shown in \figref{fig:comp_mask}.
This scheme allows more features representing the background area
(relative to the salient instance) to be viewed by the segmentation branch,
enlarging the receptive field of the segmentation branch and
meanwhile enhancing the contrast between foreground and background,
especially when there are occlusions.


\renewcommand{\addFig}[1]{\includegraphics[height=0.172\linewidth]{feature_maps/#1}}
\begin{figure*}[t]
  \centering
  \small
  \renewcommand{\tabcolsep}{1pt}
  \begin{tabular}{cccc}
    \addFig{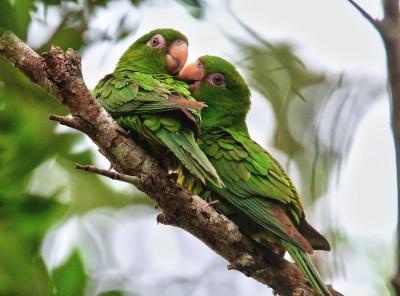} &
    \addFig{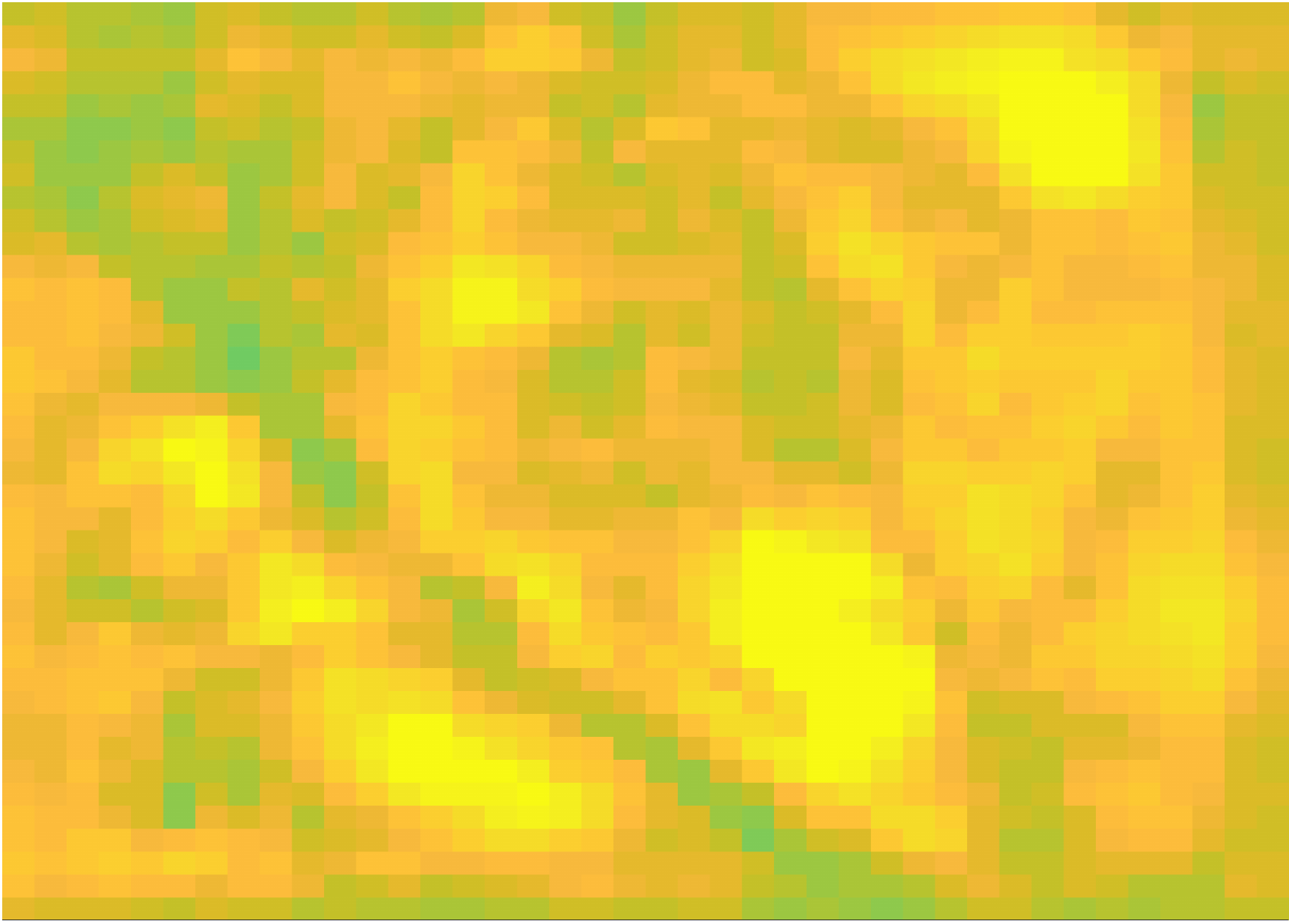} &
    \addFig{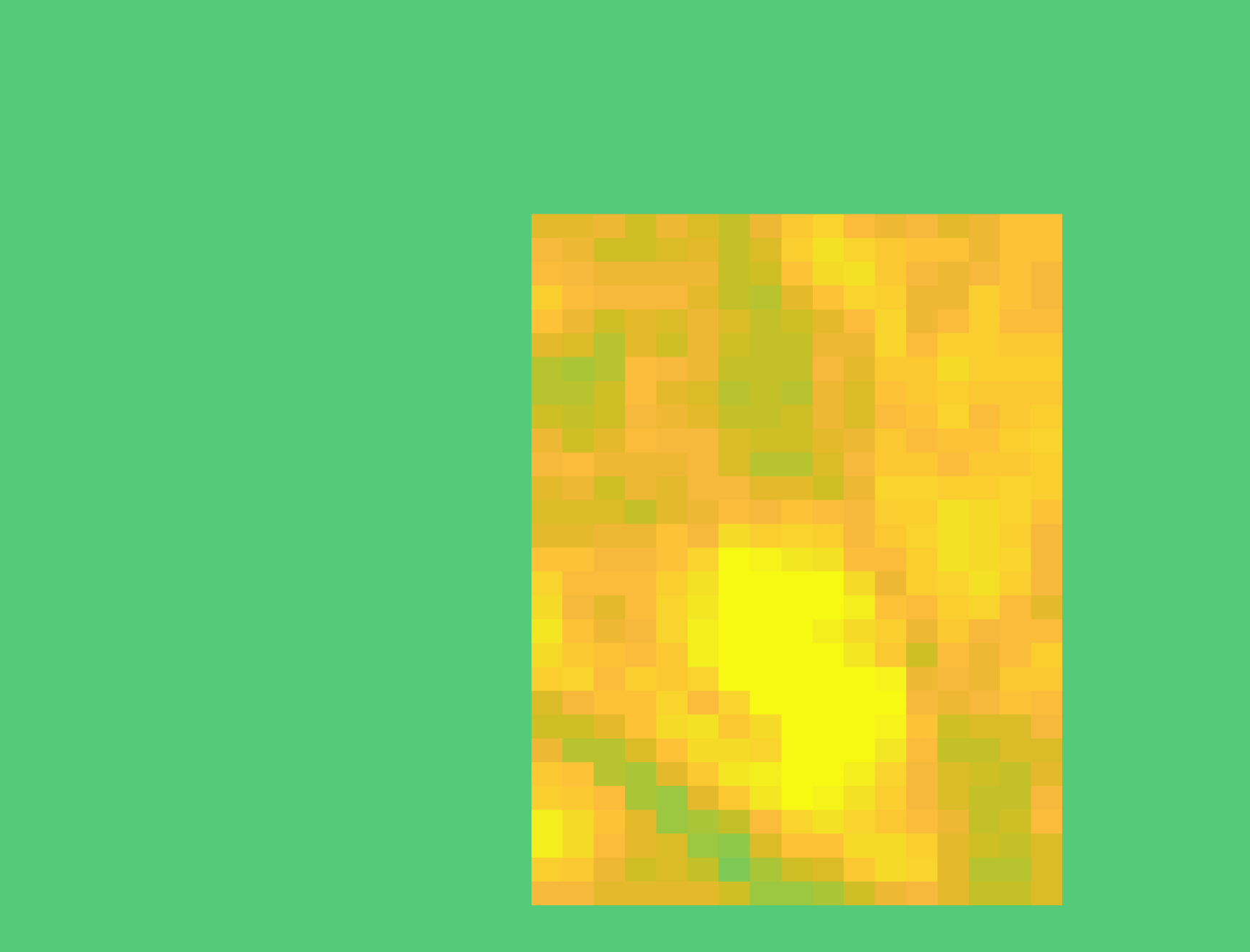} &
    \addFig{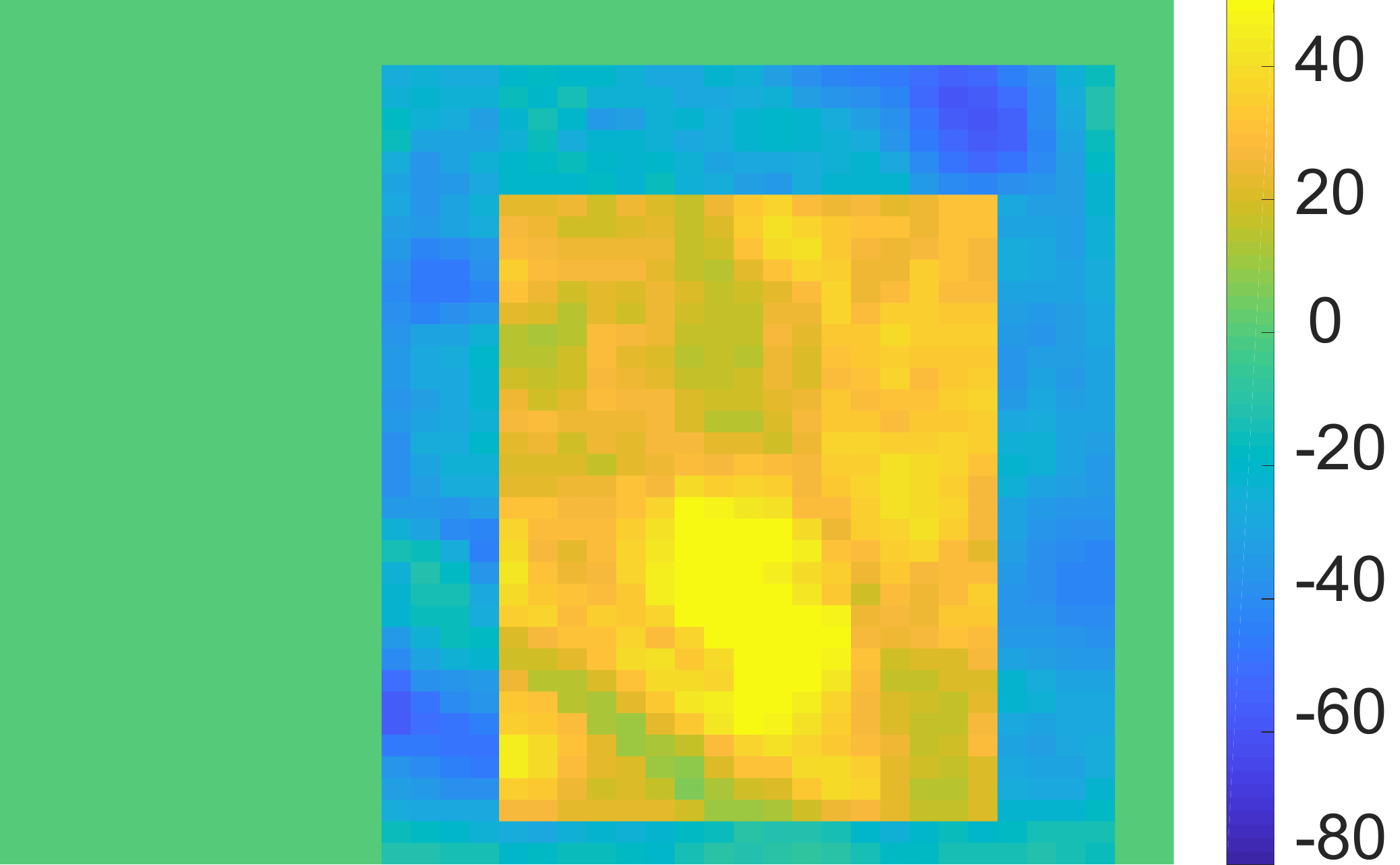} \\
    (a) Input image & (b) Feature map before masking &
    (c) Binary RoIMasking & (d) Ternary RoIMasking   \\
  \end{tabular}
  \CaptSpace
  \caption{The output feature maps of two different types of RoIMasking layers. 
  	(b) Before RoIMasking, all of the values in the feature map are 
  	non-negative because of the ReLU layer. 
  	(c) After binary RoIMasking, the regions outside the proposal are set to zeros. 
  	(d) Ternary RoIMasking additionally considers a larger area, 
  	in which feature values are non-positive.
  }\label{fig:feature_maps}
\end{figure*}

\subsection{Framework}

The pipeline of S4Net is shown in \figref{fig:arch},
which involves two components:
a bounding box detector and a segmentation branch.
Both components share the same base model.
As in most object detection works,
we select ResNet-50 \cite{He2016} as our base model.
%
%

%

\myPara{Single-Shot Object Detector.}
Considering the efficiency of the entire network,
we adopt a single-shot object detector \cite{lin2017focal}
with FPN \cite{lin2017feature} as the base model
in order to leverage the multi-level features.
%
%
To reduce the runtime cost,
we discard the lateral connected to conv2 while keeping the rest unchanged
(\ie conv3-conv6).
Four detection heads are connected to each lateral layer as shown in
\figref{fig:arch}(a).
The head structure is the same to the one used in Faster R-CNN \cite{ren2017faster},
but with different strides to perform detection at multiple scales.
%
%

\myPara{Single-Shot Segmentation Branch.}
Different from existing instance level semantic segmentation methods,
such as Mask R-CNN \cite{he2017mask},
our segmentation branch is also single-shot.
The bounding boxes predicted by the detection branch
and the output of the lateral layer with stride 8 in the backbone network
are fed into our segmentation branch.
As shown in \figref{fig:arch}(a), our segmentation branch contains
a RoIMasking layer for instance feature extraction
and a salient instance discriminator for identifying salient instances.
%

\renewcommand{\addFig}[1]{\includegraphics[width=0.323\linewidth]{#1}}
\begin{figure}[t!]
  \centering
  \small
  \renewcommand{\tabcolsep}{1pt}
  \begin{tabular}{ccc}
    \addFig{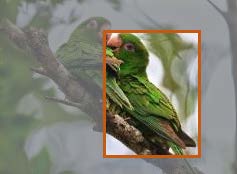} &
    \addFig{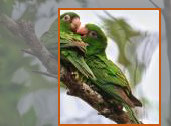} &
    \addFig{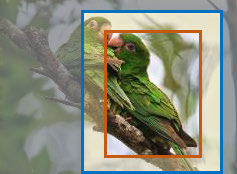} \\
    (a) Binary mask & (b) Ex-Binary mask & (c) Ternary mask \\
  \end{tabular}
  \CaptSpace
  \caption{Three different types of masks used in our RoIMasking layer. 
  	(a) Binary mask only considers the regions inside the orange rectangle;
    (b) Expanded binary mask considers a larger rectangle area than binary masking;
  	(b) The ternary mask takes into account both the region inside 
  	the orange rectangle and its surrounding regions marked in yellow.
  }\label{fig:comp_mask}
\end{figure}

\subsection{RoIMasking}\label{sec:RoIMasking}



RoIPool \cite{fastRcnn15} and RoIAlign \cite{he2017mask}
are two standard operations for extracting fixed-size features
from the regions of interest.
%
%
Both RoIPool and RoIAlign sample a region of interest into
a fixed spatial extent of $H \times W$, and typically $H = W$,
\eg $7\times 7$ in \cite{fastRcnn15} and $28\times 28$ in \cite{he2017mask}.
The RoIPool first quantizes RoIs by
uniformly dividing them into $H \times W$ spatial bins.
After max-pooling each spatial bin,
the output feature maps with size $H \times W$ can be generated.
Since the quantization in RoIPool is performed by rounding operation,
it introduces misalignments between the RoI and the extracted features.
As a remedy, RoIAlign avoids quantization by using bilinear interpolation.

However, both RoIPool and RoIAlign focus on the regions inside the proposals, 
neglecting the rest region.
As discussed in \secref{sec:observation},
the region surrounding the current object RoI contains valuable information
for distinguishing between the target object and its background.
%
Unfortunately, although some layer-fusion techniques 
such as feature pyramid network~\cite{lin2017feature} attempt to embed high-level and comprehensive information in a feature map,
both RoIPool and RoIAlign do not explicitly and effectively explore the information surrounding the RoI.
Moreover, the sampling process in these two operations makes these operations
unable to maintain the aspect ratio and resolution of the regions of interest,
possibly hurting the quality of the results.
In this subsection, we design a new resolution-preserving and 
quantization-free layer, called \textit{RoIMasking}, 
to take the place of RoIPool or RoIAlign.
We also attempt to explore feature separation between foreground and background
regions for improving segmentation quality.



\myPara{Binary RoIMasking.}
We first introduce a simplified version of RoIMasking which we call 
binary RoIMasking.
The binary RoIMasking receives feature maps and proposals predicted 
by the detection branch.
A binary mask is generated according to the position and size of a given rectangle proposal.
The values inside the rectangle are set to 1 and otherwise 0.
\figref{fig:comp_mask}a illustrates a binary version of RoIMasking, 
in which the bright and dark areas are associated with labels 1 and 0, respectively.
The output of the binary RoIMasking layer is the 
input feature maps multiplied by this mask.
In \figref{fig:feature_maps}, 
we show a typical example of the output feature maps.
Different from RoIPool \cite{fastRcnn15} and RoIAlign \cite{he2017mask}, 
our binary RoIMasking keeps the original aspect ratio and resolution 
of the feature maps.
In \secref{sec:experiments}, 
we experimentally verify that the proposed binary RoIMasking outperforms 
the RoIPool and RoIAlign baselines.

\myPara{Expanded Binary RoIMasking.} 
In this paragraph, we also consider
an extensive version of binary RoIMasking by simply enlarging the proposal region
as illustrated in \figref{fig:comp_mask}b.
Compared to the standard binary RoIMasking, expanded binary RoIMasking
takes into account more background/context information, 
which means the segmentation branch has a larger receptive field.
We will show more quantitative comparisons  in our experiment section.

\myPara{Ternary RoIMasking.}
To make better use of the background information around the regions of interest, 
we further advance the expanded binary RoIMasking to a ternary case.
Because of the ReLU activation function, 
there are no negative values in the feature maps before RoIMasking.
To explicitly notify the segmentation branch that the region outside
the proposals should be considered as background,
we flip the signs (\ie set the corresponding mask values to -1) of the feature values 
around the region of interest,
which is illustrated in yellow color in \figref{fig:comp_mask}c.
In this way, the features around regions of interest are distinct 
from those inside the bounding boxes of the salient instances.
This allows the segmentation branch to be able to not only 
make use of features inside the region of interest as well 
as the surrounding context (as in extended binary RoIMasking),
but also explicitly emphasis on foreground/background feature separation. 
%
The feature map after ternary RoIMasking is illustrated in \figref{fig:feature_maps}d.
It is worth mentioning that this operation introduces no additional 
computation cost into our model.
Ternary RoIMasking leads to a large improvement as we show in the experiment part
(\secref{sec:experiments}).
In the following, we \emph{abbreviate ternary RoIMasking as RoIMasking} 
for notational convenience unless otherwise noted.

\subsection{Analysis of RoIMasking}

\renewcommand{\addFig}[1]{\includegraphics[height=0.31\linewidth]{gradients/#1}}
\begin{figure}[t]
  \centering
  \renewcommand{\tabcolsep}{1pt}
  \begin{tabular}{cc}
    \addFig{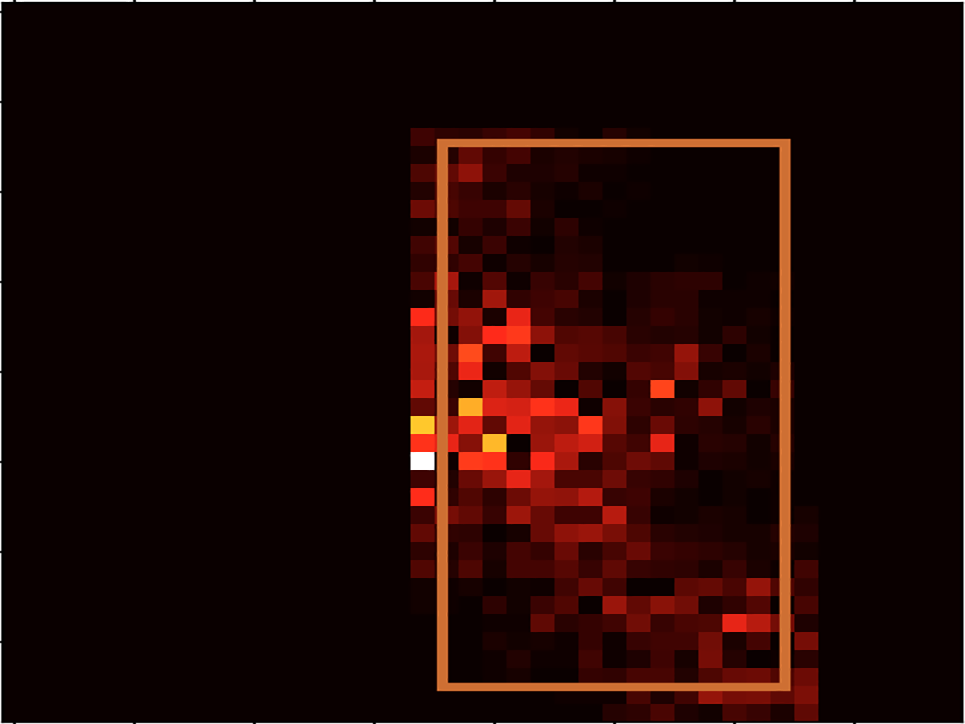} &
    \addFig{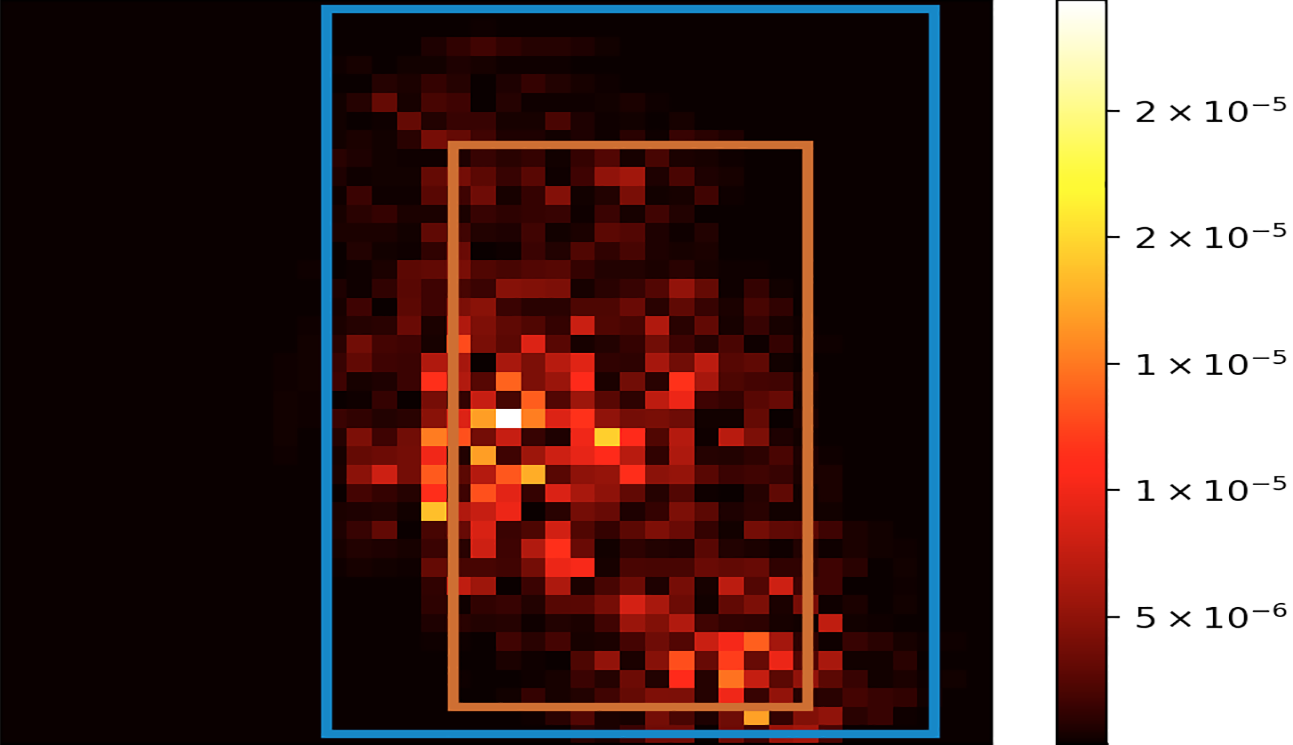} \\
    (a) Binary masking & (b) Ternary masking \\
  \end{tabular}
  \CaptSpace
  \caption{Gradient maps using binary masking and ternary masking. 
  	As can be seen, ternary masking considers more perimeter information 
  	of the region around the proposal. 
  	The input image in this experiment is shown in \figref{fig:feature_maps}(a).
  }\label{fig:Gradient_maps}
\end{figure}

This subsection demonstrates the importance of the background information around 
the regions of interest in the feature maps and the effectiveness of ternary RoIMasking.
To do so, we explore the impact of each activation in the feature maps 
before RoIMasking on the performance.
Inspired by \cite{yosinski2015understanding}, 
we visualize the function of a specific neuron in this model by drawing a gradients map.
After loading the fully trained model weights, 
we do a forward pass using a specific image.
In this process, the activation value of the feature maps before 
RoIMasking, $H_{i,j,c}$, is extracted and stored.
Next, we do a backward pass.
Note that in the general training stage, 
back-propagation is performed to calculate the gradients of the 
\emph{total loss} with respect to the \emph{weights} in the neural network.
But in this experiment, we load the stored $H_{i,j,c}$ as a variable, 
and regard the convolution kernels as constant.
Back-propagation is performed to calculate the gradients of the 
\emph{instance segmentation loss} with respect to each \emph{feature map} input to RoIMasking,
\ie $G_{i,j,c}={\partial L_{sal}}/{\partial H_{i,j,c}}.$
The absolute value of $G_{i,j,c}$ reflects the importance of the feature 
map pixel $H_{i,j,c}$ to the saliency task.
After summing up $|G_{i,j,c}|$ along the channel dimension, 
the gradient map $G_{i,j}$ can be obtained.

\figref{fig:Gradient_maps} shows the gradient maps for binary RoIMasking 
and ternary RoIMasking, respectively.
The orange rectangle is the ground truth bounding box of a salient instance.
By definition, the pixels inside the orange rectangle in the ternary mask 
are set to 0 and the pixels between the orange and blue boxes are set to -1.
It is obvious that there are evident responses in the background 
(marked as `-1' in the ternary mask) area in \figref{fig:Gradient_maps}b.
In \figref{fig:Gradient_maps}a, 
there are only few responses between the orange and blue boxes.
%
This phenomenon indirectly indicates the importance of the context information 
around the regions of interest.
%
More experimental results can be found in the experiment section.
 
\subsection{Segmentation Branch}

%
Taking into account the structure of our backbone,
we take the feature maps from the lateral layer associated with
conv3 with a stride of 8 as the input to our segmentation branch 
on the trade-off between global context and details.
Before connecting our RoIMasking layer, 
we first add a simple convolutional layer with 256 channels and
kernel size $1 \times 1$ for compressing the number of channels.
%
Despite the RoIMasking layer, it is still difficult to distinguish the salient instances 
from the other instances inside the same RoI.
%
To this end, we add a segmentation branch similar to Mask-RCNN \cite{he2017mask}
to help better distinguish the instances.

As pointed out in \cite{zhao2016pyramid}, 
enlarging receptive field is helpful for segmentation related tasks.
Inspired by \cite{zhao2016pyramid,he2017mask}, we design a new segmentation branch
by introducing skip connections and dilated convolutional layers (See \figref{fig:arch}c).
Other than two residual blocks, 
we also add two $3 \times 3$ max pooling with stride 1 and 
dilated convolutional layers with dilation rate 2 for enlarging the receptive field.
All the convolutional layers have a kernel size $3 \times 3$ and stride 1.
For the channel numbers, we set the first three to 128 and the rest 64,
which we found are enough for salient instance segmentation.

%

\subsection{Loss function}
As described above, there are two sibling branches in our framework for 
detection and saliency segmentation, respectively.
The detection branch undertakes objectness classification task and coordinates 
regression task, and the segmentation branch is for saliency segmentation task.
Therefore, we use a multi-task loss $L$ on each training sample to jointly train 
the model:
\begin{equation}
L = L_{obj} + L_{coord} + L_{seg}.
\end{equation}
Regarding the fact that positive proposals are far less than negative samples in the detection branch,
we adopt the following strategy.
Let $P$ and $N$ be the collections of positive and negative proposals, $N_P$ and $N_N$ be the numbers of positive and negative proposals ($N_P \ll N_N$), then we
calculate the positive and negative objectness loss separately to avoid the domination of negative gradients
during training.
Thus we have:
\begin{equation} \label{eqn:pos_neg}
L_{obj} = -( \frac{1}{N_P} \sum_{i \in P}^{} \log p_i + \frac{1}{N_N} \sum_{j \in N}^{} \log (1 - p_j)),
\end{equation}
in which $p_i$ is the probability of the $i$th proposal being positive.

We use $Smooth_{L1}$ loss as Fast-RCNN \cite{fastRcnn15} for coordinate regression and cross-entropy loss similar to Mask-RCNN \cite{he2017mask}
for the segmentation branch.



\section{Experiments}\label{sec:experiments}

In this section, we carry out detailed analysis to elaborate the functions of each component in
our method by ablation studies.
We also perform thorough comparisons with the state-of-the-art methods to 
exhibit the effectiveness of our approach.
We use the dataset proposed in \cite{li2017instance} for all experiments.
This dataset contains 1,000 images with well-annotated instance-level annotations.
For fair comparisons, as done in \cite{li2017instance}, we randomly select 500 images for training,
200 for validation, and 300 for testing.
%

\newcommand{\thickhline}{%
    \noalign {\ifnum 0=`}\fi \hrule height 1pt
    \futurelet \reserved@a \@xhline
}

\newlength\savedwidth
\newcommand{\whline}[1]{\noalign{\global\savedwidth\arrayrulewidth \global\arrayrulewidth #1}%
                   \hline \noalign{\global\arrayrulewidth\savedwidth}}

\newcolumntype{"}{@{\hskip\tabcolsep\vrule width 0.8pt\hskip\tabcolsep}}

\newcommand{\RoIAlign}{RoIAlign (Mask R-CNN  \cite{he2017mask})}
\newcommand{\RoIPool}{RoIPool (Fast R-CNN \cite{fastRcnn15}.)}


\renewcommand{\arraystretch}{1.2}
\begin{table}[htp]
  \centering
  \small
  \renewcommand{\tabcolsep}{4pt}
  \begin{tabular}{c|cc|cc} 
  \toprule[1pt]
    Methods & $\text{mAP}^{0.5}$ & $\text{mAP}^{0.7}$ & $\text{mAP}^{0.5}_O$ & $\text{mAP}^{0.7}_O$\\ \midrule[1pt]
    RoIAlign \cite{he2017mask} &  $85.2\%$ & $61.5\%$ & $79.2\%$ & $47.7\%$ \\
    RoIPool \cite{fastRcnn15} &  $85.2\%$ & $61.1\%$ & $80.3\%$ & $50.9\%$ \\ \midrule[1pt]
    Binary RoIMasking  	  &  $85.5\%$ & $62.4\%$ & $80.1\%$ & $49.4\%$ \\
    Ternary RoIMasking     &  $86.7\%$ & $63.6\%$ & $81.2\%$ & $51.5\%$
    \\ 
    \bottomrule[1pt]
  \end{tabular}
  \caption{Ablation experiments for analyzing our RoIMasking layer.
  We also list the results using the RoIAlign and RoIPool proposed in Mask R-CNN \cite{he2017mask}
  and Fast R-CNN \cite{fastRcnn15}, respectively.
  %
  %
  Obviously, our proposed RoIMasking outperforms RoIAlign and RoIPool even for the images
  with occlusion.
  }\label{table:validation_study}
\end{table}

\subsection{Implementation Details}

\myPara{Training and Testing.}
In the training phase, the IoU is used to determine whether a bounding box 
proposal is a positive or negative sample in the detection branch.
%
A bounding box proposal is positive if it's IoU $> 0.5$, and negative if IoU $<$ 0.5.

In the testing phase, the bounding boxes fed into 
the RoIMasking layer are 
from the detection branch.
But in the training phase, 
we directly feed the ground truth bounding boxes into the RoIMasking layer.
This provides the segmentation branch with more stable and 
valid training data and meanwhile accelerates the training process,
as been verified by empirical experiments.

\myPara{Hyper-parameters.}
Our proposed network is based on the
TensorFlow library \cite{abadi2016tensorflow}.
The input images are augmented by horizontal flipping.
The hyper-parameters are set as follows: weight decay (0.0001) and momentum (0.9).
We train our network on 2 GPUs for 20k iterations, 
with an initial learning rate of 0.004 which is divided by
a factor of 10 after 10k iterations.
%

\renewcommand{\AddFig}[1]{\includegraphics[height=0.162\linewidth]{Demos/#1.jpg}}
\begin{figure*}[t]
  \centering
  \renewcommand{\tabcolsep}{1pt}
  \AddFig{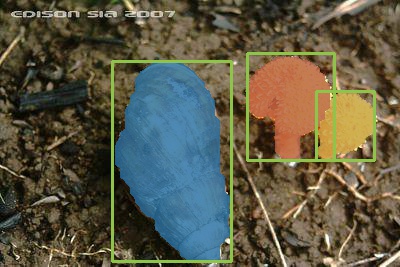}
  \AddFig{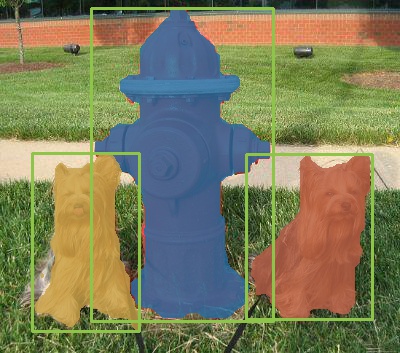}
  \AddFig{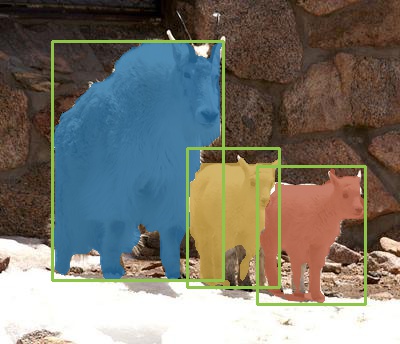}
  \AddFig{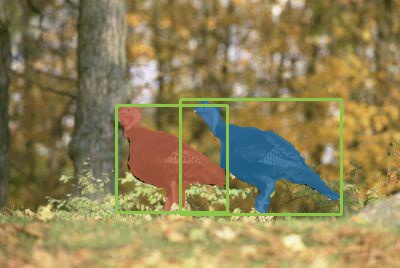}
  \AddFig{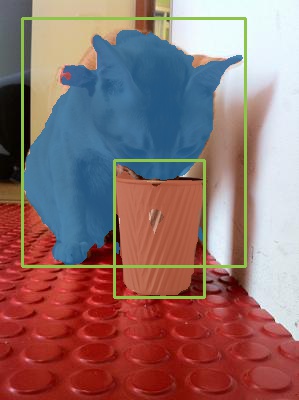} \\ 
  \CaptSpace
  \caption{Selected examples of instance-level saliency segmentation
  	results on the dataset proposed by \cite{li2017instance}
  	(the line above) and COCO \cite{lin2014microsoft} (the line below).
  	Even obstructed instances can be well distinguished and segmented
  	by our \S4Net.
  } \label{fig:Examples}
\end{figure*}

\subsection{Ablation Studies} \label{ablation_study}


To evaluate the effectiveness of each component in our proposed framework, 
we train our model on the salient instance segmentation dataset \cite{li2017instance}.
Following the standard COCO metrics \cite{lin2014microsoft}, we report results on
mAP (averaged precision over IoU thresholds), including
$\text{mAP}^{0.5}$ and $\text{mAP}^{0.7}$.
%
Furthermore, to analyze the ability to distinguish different instances, 
we also consider another set which only contains instances with occlusion, 
which is  denoted by $\text{mAP}_{O}$.

\myPara{Effect of RoIMasking.}
To evaluate the performance of the proposed RoIMasking layer, 
we also consider using RoIPool \cite{fastRcnn15} and RoIAlign \cite{he2017mask}.
We replace our RoIMasking with RoIPool and RoIAlign to 
perform two comparison experiments while
keep all other network structures and experimental settings unchanged.
Quantitative evaluation results are listed in \tabref{table:validation_study}.
As can be seen, our proposed binary RoIMasking and ternary RoIMasking 
both outperform RoIPool and RoIAlign in $\text{mAP}^{0.7}$.
Specifically, our ternary RoIMasking result improves the result using RoIAlign 
by around 2.1 points.
This reflects that considering more context information outside
 the proposals does help
for salient instance segmentation.

To further verify the effectiveness of our RoIMasking layer, we also consider a binary
masking case in which the values in the yellow area of \figref{fig:comp_mask}d are all
set to 1.
The penultimate line in Table~\ref{table:validation_study} shows the corresponding results.
As can be seen, the results are even worse when the binary masking is used.
This fact reflects that simply enlarging the regions of interest is not 
helpful for discriminating salient instances.
However, when the signs of the extended regions in the mask are flipped (ternary RoIMasking),
the best results can be obtained (the bottom line of Table~\ref{table:validation_study}).
This demonstrates that changing the signs of the extended regions in the mask
can explicitly increase the contrast between the salient instances and background.
More importantly, the non-salient regions inside the proposals will tend to be
predicted to the same class as the extended regions.
Therefore, the feature separation ability between the target objects 
and their nearby background plays an important role in our approach.
%


\begin{table}[htp]
    \centering
    \small
     \begin{tabular}{x{16mm}|x{10mm} x{10mm} x{10mm} x{10mm} x{10mm} x{10mm}} 
          \toprule[1pt]
          $\alpha$ & $0$ & $1/6$ & $1/3$ & $1/2$ & $2/3$ & $1$ \\   \midrule[1pt]
          $\text{mAP}^{0.5}$ & $85.9\%$ & $86.4\%$ & $\mathbf{86.7\%}$ & $86.5\%$ & $86.2\%$ & $85.9\%$ \\
          $\text{mAP}^{0.7}$ & $62.5\%$ & $63.4\%$ & $\mathbf{63.6\%}$ & $63.3\%$ & $62.4\%$ & $62.0\%$ \\
          \bottomrule[1pt]
        \end{tabular}
        \caption[]{Performance of S4Net with different expanded areas. All the results shown here
        are based on ResNet-50 \cite{He2016}. As can be observed, when $\alpha=1/3$ we obtain the best result.}
        \label{table:alpha}
\end{table}

\myPara{Size of Context Regions.}
To better understand our RoIMasking layer, 
we analyze how large the context regions should be here.
Suppose the bounding box size of a salient instance is $(w, h)$,
where $w$ and $h$ are the width and height, respectively.
We define an expansion coefficient $\alpha$ 
to denote the width of the `-1' region in the RoI mask.
So, the size of the valid region is $(w+2{\alpha}w, h+2{\alpha}h)$.
By default, we set $\alpha$ to 1/3.
We also try different values of $\alpha$ to explore its influence on the final 
results as shown in Table \ref{table:alpha} but found
both larger and smaller values of $\alpha$ slightly harms the performance.
This indicates that a region size of $(w+2w/3, h+2h/3)$ 
is enough for discriminating different instances as larger `-1' region
may make more salient instances be viewed, weakening the performance of
identifying the `real' salient instances.

\myPara{Number of Proposals.}
The number of proposals sent to the segmentation branch also effects the performance.
According to our experiments, 
more proposals lead to better performance but more computational costs.
%
%
Notice that the performance gain is not obvious when the number of proposals exceeds 20.
Specifically, when we set the number of proposals to 100, 
only around 1.5\% improvement can be achieved
but the runtime cost increases dramatically.
Taking this into account, 
we take 20 proposals as a trade-off during the inference phase.
Users may decide the number of proposals by their tailored tasks.

\begin{table}[htp]
    \centering
    \small
    \begin{tabular}{x{30mm}"x{15mm} x{15mm}"x{18mm}} 
     \toprule[1pt]
          Base models & mAP@0.5 & mAP@0.7 & Speed (FPS)\\ \midrule[1pt]
          ResNet-101 \cite{He2016} &  $\mathbf{88.1\%}$ & $\mathbf{66.8\%}$ & $33.3$\\
          ResNet-50 \cite{He2016} &  $86.7\%$ & $63.6\%$ & $40.0$\\ \midrule[1pt]
         VGG16 \cite{simonyan2014very} & $82.2\%$ & $53.0\%$ & $43.5$\\
         MobileNet \cite{howard2017mobilenets} & $62.9\%$ & $33.5\%$ & $90.9$\\ 
          \bottomrule[1pt]
        \end{tabular}
        \caption[]{Performance of \S4Net when using different base models. When we change the default ResNet-50
        to ResNet-101, another 3.2\% improvement can be obtained in spite of a little sacrifice on time cost.
        We also attempt to use the recent MobileNet \cite{howard2017mobilenets} as our base model and yield
        a frame rate of more than 90 fps on a GTX 1080 Ti GPU.}
        \label{table:basemodel_exp}
\end{table}

\myPara{Base Models.} Besides the base model of ResNet-50 \cite{He2016}, 
we also try another three popular base models, 
including Resnet-101 \cite{He2016}, VGG16 \cite{simonyan2014very},
and MobileNet \cite{howard2017mobilenets}.
\tabref{table:basemodel_exp} lists the results when different 
base models are utilized.
As one can see, base models with better performance on classification 
also works better in our experiments.
For speed, real-time processing can be achieved by our proposed S4Net.
When the size of input images is $320 \times 320$, 
S4Net has a frame rate of 40.0 fps on a GTX 1080 Ti GPU.
Furthermore, using MobileNet \cite{howard2017mobilenets} as our base model, 
S4Net runs very fast at a speed of 90.9 fps.

\newcommand{\FCIS}{FCIS \cite{li2016fully}}
\newcommand{\MSRNet}{MSRNet \cite{li2017instance}}
\newcommand{\MkRCNN}{Mask R-CNN \cite{he2017mask}}

\begin{table}[tp]
    \centering
    \small
    \begin{tabular}{x{22mm}"x{12mm} x{12mm}"x{12mm} x{12mm}}
        \toprule[1pt]
        Methods & $\text{mAP}^{0.5}$ & $\text{mAP}^{0.7}$ & $\text{mAP}^{0.5}_O$ & $\text{mAP}^{0.7}_O$ \\ \midrule[1pt]
        \MSRNet & $65.3\%$ & $52.3\%$ & - & -   \\
        S4Net &  $\mathbf{86.7\%}$ & $\mathbf{63.6\%}$ & $81.2\%$ & $51.5\%$ \\
        \bottomrule[1pt]
    \end{tabular}
    \caption[]{Quantitative comparisons with existing methods on the `test' set.
    As the instance segmentation maps of \cite{li2017instance} and 
    related code are not available, thus we use `-' to fill the blank cells.}
    \label{table:comp_exp_old}
    \vspace{-8pt}
\end{table}


\subsection{Comparisons with the State-of-the-Arts}

Unlike salient object detection which has been studied for years,
salient instance detection is a relatively new problem such that  
there is only one related work MSRNet \cite{li2017instance}
that can be used for direct comparison.
%
%
In this experiment, we compare our S4Net based on ResNet-50 \cite{He2016}
with the MSRNet method.
We report the results on the `test' set using $\text{mAP}^{0.5}$ and $\text{mAP}^{0.7}$ metric.
%

\myPara{Quantitative Analysis.}
Two datasets are used in our comparison experiments.
The results of comparative experiments on dataset proposed
by \cite{li2017instance} are listed in \tabref{table:comp_exp_old}.
Our proposed S4Net achieves better results in both 
$\text{mAP}^{0.5}$ and $\text{mAP}^{0.7}$ compared to MSRNet \cite{li2017instance}.
Specifically, our approach improves the baseline results presented in 
MSRNet \cite{li2017instance} by about 21 points in $\text{mAP}^{0.5}$.
Regarding $\text{mAP}^{0.7}$, 
we also have a great improvement on the same dataset.
%
%
%

\section{Applications}

In this section, we apply our proposed S4Net to a popular vision task---weakly-supervised
semantic segmentation.
For training samples with multiple keywords,
such as the images in PASCAL VOC~\cite{everingham2015pascal},
discriminating different instances is even essential for keyword assignment.
The detailed methodology can be seen in our supplementary material.
The results is shown in Table \ref{tab:comps_weakly}.
It is obvious that training with our instance segmentation on the same dataset
works much better than the settings in which other heuristic cues are used.
Our approach obtains a 4.8\% performance gain compared to using
the DSS salient object detector \cite{hou2016deeply}.

\begin{table}[t!]
  \centering
  \small
  \begin{tabular}{x{32mm}"x{32mm}"x{15mm}}
  \toprule[1pt]
     Model & Heuristic cues & val set \\ \midrule[1pt]
     DeepLab-VGG16 & Sal maps \cite{jiang2013salient} & $49.8\%$ \\
     DeepLab-VGG16 & Sal maps \cite{hou2016deeply} & $52.6\%$ \\
     DeepLab-VGG16 & Att \cite{zhang2016top} + Sal \cite{hou2016deeply} & $53.8\%$ \\ \midrule[1pt]
     DeepLab-VGG16    & Salient Instances       & $57.4\%$ \\
     DeepLab-ResNet101    & Salient Instances       & $\mathbf{61.8\%}$ \\ 
     \bottomrule[1pt]
  \end{tabular}
  \caption{Semantic segmentation results with different initial heuristic cues on the 
  PASCAL VOC validation set. 
  The best result is highlighted in \textbf{bold}. 
  Due to the space limitation, we use abbreviations for convenience. 
  As can be seen, training with our instance-level saliency cues greatly 
  outperforms settings with regular saliency cues.
  }
  \label{tab:comps_weakly}
\end{table}

\section{Conclusions}
In this paper, we present a single stage salient-instance segmentation framework, 
which is able to segment instance-level salient objects in real time.
The key novelties include (i) the ROIMasking layer,
which takes into
account both the information inside the proposals
and the context information outside the proposals
and preserves the original resolution and aspect ratio of 
the regions of interest,
and (2) an advanced salient instance discriminator
which enlarges the receptive field
of our segmentation branch and thus
boosts the performance.
Thorough experiments show that the proposed RoIMasking greatly outperforms RoIAlign and RoIPool, 
especially for distinguishing instances in the same scope.
Our S4Net achieves the state-of-the-art performance on a publicly available benchmark.

\section*{Acknowledgements}
This research was supported by NSFC (61521002, 61572264, 61620106008), 
the Joint NSFC-ISF Research Program (project number 61561146393), Tsinghua-Tencent Joint Laboratory for Internet Innovation Technology,
the national youth talent support program,
Tianjin Natural Science Foundation (17JCJQJC43700, 18ZXZNGX00110)
and the Fundamental Research Funds for the Central Universities 
(Nankai University, NO. 63191501).
Shi-Min Hu is the correcponding author of the paper.

{\small
\bibliographystyle{ieee}
\bibliography{S4Net}
}

\end{document}